\documentclass{article}
\usepackage{spconf,amsmath,graphicx}
\usepackage{amssymb}
\usepackage{xcolor}
\usepackage{enumitem}
\usepackage{float}
\usepackage{multirow}
\usepackage{booktabs}

\def \bR {\mathbb{R}}

\newcommand{\atj}{}

\title{Improved Language Identification \\Through Cross-Lingual Self-Supervised Learning}
\name{\begin{tabular}{c}Andros Tjandra, Diptanu Gon Choudhury, Frank Zhang, Kritika Singh,\\ Alexis Conneau\sthanks{Currently at Google AI, work done while at Facebook AI.}, Alexei Baevski, Assaf Sela, Yatharth Saraf, Michael Auli \end{tabular}}
\address{Facebook AI, USA \\\texttt{\small{\{androstj,diptanu\}@fb.com}}}
\begin{document}
\ninept
\maketitle
\begin{abstract}
Language identification greatly impacts the success of downstream tasks such as automatic speech recognition.
Recently, self-supervised speech representations learned by wav2vec 2.0 have been shown to be very effective for a range of speech tasks.
We extend previous self-supervised work on language identification by experimenting with pre-trained models which were learned on real-world unconstrained speech in multiple languages and not just on English.
We show that models pre-trained on many languages perform better and enable language identification systems that require very little labeled data to perform well.
Results on a 26 languages setup show that with only 10 minutes of labeled data per language, a cross-lingually pre-trained model can achieve over 89.2\% accuracy.
\end{abstract}
\section{Introduction}
\label{sec:intro}
Automatic speech recognition (ASR) has seen large improvements through better modeling~\cite{linhao2018transformer,gulati2020conformer,wang2020hybrid} and the use of unlabeled data~\cite{synnaeve2020end,xu2020iterative,park2020improved,chung2018speech2vec,baevski2020wav2vec}.
Despite a sizeable body of work on multilingual speech recognition~\cite{burget2010multilingual, lin2009study,heigold2013multilingual,bourlard2011current,cho2018multilingual,toshniwal2018multilingual,Kannan_2019,li2019bytes,pratap2020massively}, the  vast majority of systems are trained for a single language.
However, in many real-world settings, we wish to transcribe speech data in different languages and it is crucial to route utterances to the system trained for the language at hand.
Language identification (LID) is typically used to identify the language of an utterance and the accuracy of this component is crucial to prevent poor ASR performance.

Language identification has been tackled with conventional methods~\cite{zissman1996comparison} as well as with modern neural networks~\cite{lopez2014lid}. 
Most of these approaches are trained purely with labeled data, however, unlabeled data is typically much easier to collect. 
Self-supervised learning leverages unlabeled data to obtain good data representations that can then be fine-tuned for a particular downstream task~\cite{oord2018cpc,chung2018speech2vec,schneider2019wav2vec,baevski2020wav2vec}.

Prior work on LID has explored the use of a wav2vec 2.0 model pre-trained only on English data~\cite{fan2020exploring}. 
In this paper, we extend this work by considering cross-lingually trained self-supervised models~\cite{conneau2020unsupervised}.
In particular, we pre-train models with a large amount of unlabeled data from many different languages and then fine-tune them with as little as 10 minutes of labeled data per language for LID to enable systems for low-resource languages. The audio data used here is sampled from public social media videos, which presents unique challenges such as a variety of speaker styles and the quality of the recordings.
Moreover, our approach does not use any auxiliary features as are commonly used to improve performance.
We also investigate different pooling strategies to aggregate the pre-trained context-representations for LID classification task.
We modified wav2vec 2.0 to use log-mel spectrogram features as input, similar to~\cite{zhang2020pushing}.
Our experiments show strong performance using as little as ten minutes of labeled data per language compared to models trained on much larger amounts of labeled data.
Furthermore, multilingual pre-trained models achieve better performance than models pre-trained only with a single language.

\begin{figure*}
\centering
\includegraphics[width=0.80\linewidth]{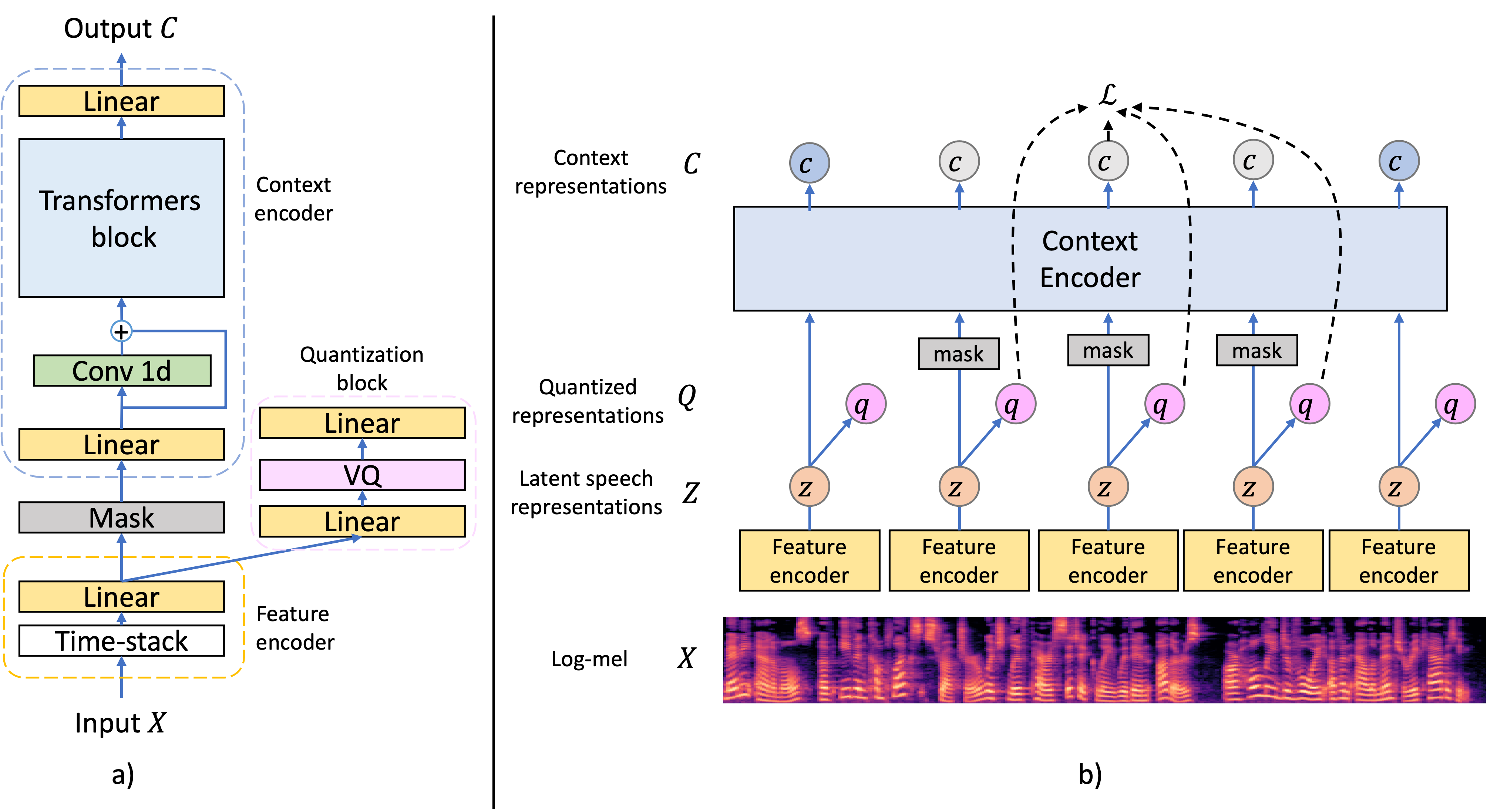}
\label{fig:logmel_wav2vec}
\caption{a) Log-mel Wav2Vec architecture. \quad b) An illustration of how Wav2Vec generates context representation and solves the contrastive task.}
\end{figure*}

\section{Log-mel Wav2Vec Architecture}

In this section, we describe our modifications to the original wav2vec 2.0 model architecture and the cross-lingual training strategy.
\atj{A wav2vec model \cite{baevski2020wav2vec, schneider2019wav2vec} consists of multiple convolution and Transformer \cite{vaswani2017attention} layers and operates on top of the raw waveform. 

Here, we use log-mel spectrogram \cite{zhang2020pushing} as the input features instead of raw waveform to improve the memory and computational efficiency. 
\atj{We define our input feature as $X \in \bR^{T \times F} = [x_1,..,x_S]$ where $S$ is the number of frames in an utterance and $F$ is the input dimension for each frame (e.g., $F=80$ for 80-dimensional log-mel spectrogram). From here, the input features are fed into a feature encoder which is followed by a context encoder.} We present our modified wav2vec architecture in Figure~\ref{fig:logmel_wav2vec}(a)}.

\subsection{Feature encoder}
A feature encoder $f(\cdot): X \xrightarrow{} Z$ takes input $X$ and outputs latent speech representations $Z = [z_1,...,z_{T}]$ where ${T=S/R}$. \atj{Compared to original wav2vec 2.0, we replace the seven convolutional layers in the encoder with a time-stacking layer and a linear layer}. Due to the quadratic cost $\mathcal{O}(n^2)$ of time and memory of Transformer layers, reducing the input length greatly improves the training and inference efficiency. The time-stacking layer is defined as a function $ts(\cdot): \bR^{S\times F} \xrightarrow{} \bR^{T \times FR}$, which concatenates $R$ consecutive frames into a single frame and reduces sequence length by a factor of $1/R$. 

\subsection{Context encoder}
A context encoder $g(\cdot): Z \xrightarrow{} C$ takes the input $Z$ produced by the feature encoder block and outputs context representations $C$. The context encoder consists of a linear layer, a layer normalization \cite{lei2016layer}, a convolution layer to encode relative position information \cite{baevski2020wav2vec}, multiple Transformers layers \cite{vaswani2017attention} and another linear layer.

\subsection{Quantization block} \label{sec:quantization}
A quantization block $h(\cdot): Z \xrightarrow{} Q$ takes the output $Z$ of the feature encoder layer and produces a quantized representation $Q$. A linear layer is added on top to the input $Z$. A product quantization \cite{jegou2010product, baevski2019vq} with $G$ groups and $V$ codebook entries $e \in \bR^{V \times d/G}$ was applied on the projected input. The result of each group is concatenated and another linear layer is applied to generate the quantized representation $Q$. A Gumbel softmax function \cite{jang17gumbel} enables discretely choosing an index based on the maximum value to be fully differentiable.

\section{Cross-lingual pre-training}
\atj{For multilingual pre-training, we collected large quantities of unlabeled speech from various languages and combined them into a single multilingual dataset to train cross-lingual speech representations (XLSR;~\cite{conneau2020unsupervised})}.
The audio data used here is sampled from public social media videos, involving unconstrained, natural speech that is
filled with background music and noise, with various speaking styles, disfluencies, accents, and un-cued speaker and language switching. This presents an interesting and challenging application of self-supervised learning that directly complements other recent work on self-supervised learning which focused on datasets based on audio-books, which is clean and focused on a single domain~\cite{schneider2019wav2vec,baevski2020wav2vec,zhang2020pushing} with a few exceptions~\cite{hsu2021robust}. \atj{Compared to~\cite{conneau2020unsupervised}, we do not upsample or downsample certain languages during training because we assume no access to the language ID in the unlabeled video dataset.}

Figure~\ref{fig:logmel_wav2vec}(b) illustrates how each block interacts with the other to solve the contrastive task. The model needs to find which sample is a true quantized latent $q_t$ from a set of $K+1$ candidates $\tilde{q} \in Q$. The false quantized latent $\tilde{q} \setminus q_t$ are uniformly sampled from any masked time-step. We define this contrastive loss as: 
\begin{align}
    \mathcal{L}_{m} &= -\log \frac{\exp (sim(c_t, q_t))}{ \sum_{\tilde{q} \sim Q} \exp(sim(c_t, \tilde{q}))} 
\end{align} where $sim(c_t,q_t)$ is cosine-similarity between context representations and quantized latent speech representations. 

We add a diversity loss to encourage the equal usage of $V$ codebook entries on each $G$ codebook (Sec.\ref{sec:quantization}). The diversity loss is designed to maximize the entropy of the averaged softmax probability over the codebook entries for each codebook group $\bar{p}_{g}$. We define the loss as:
\begin{align}
    \mathcal{L}_{d} &= \frac{1}{GV}\sum_{g=1}^{G}-\mathbb{H}(\bar{p}_{g}) = \frac{1}{GV}\sum_{g=1}^{G}\sum_{v=1}^{V} \bar{p}_{g,v} \log \bar{p}_{g,v} \\
    \text{where \,\,\,\,} & \bar{p}_{g,v} = \sum_{b=1}^{B}\sum_{t=1}^{T} \frac{p_{b,t,g,v}}{BT}
\end{align} and $\bar{p}_{b,t,g,v}$ is the softmax probability without gumbel noise on group $g$, codebook entry $v$, sample $b$, and time-step $t$. Our final loss is defined as $\mathcal{L} = \mathcal{L}_{m} + \lambda \mathcal{L}_{d}$ where $\lambda$ is a hyperparameter to control the diversity loss.

\section{LID finetuning}
\begin{figure}
\centering
\includegraphics[width=0.42\linewidth]{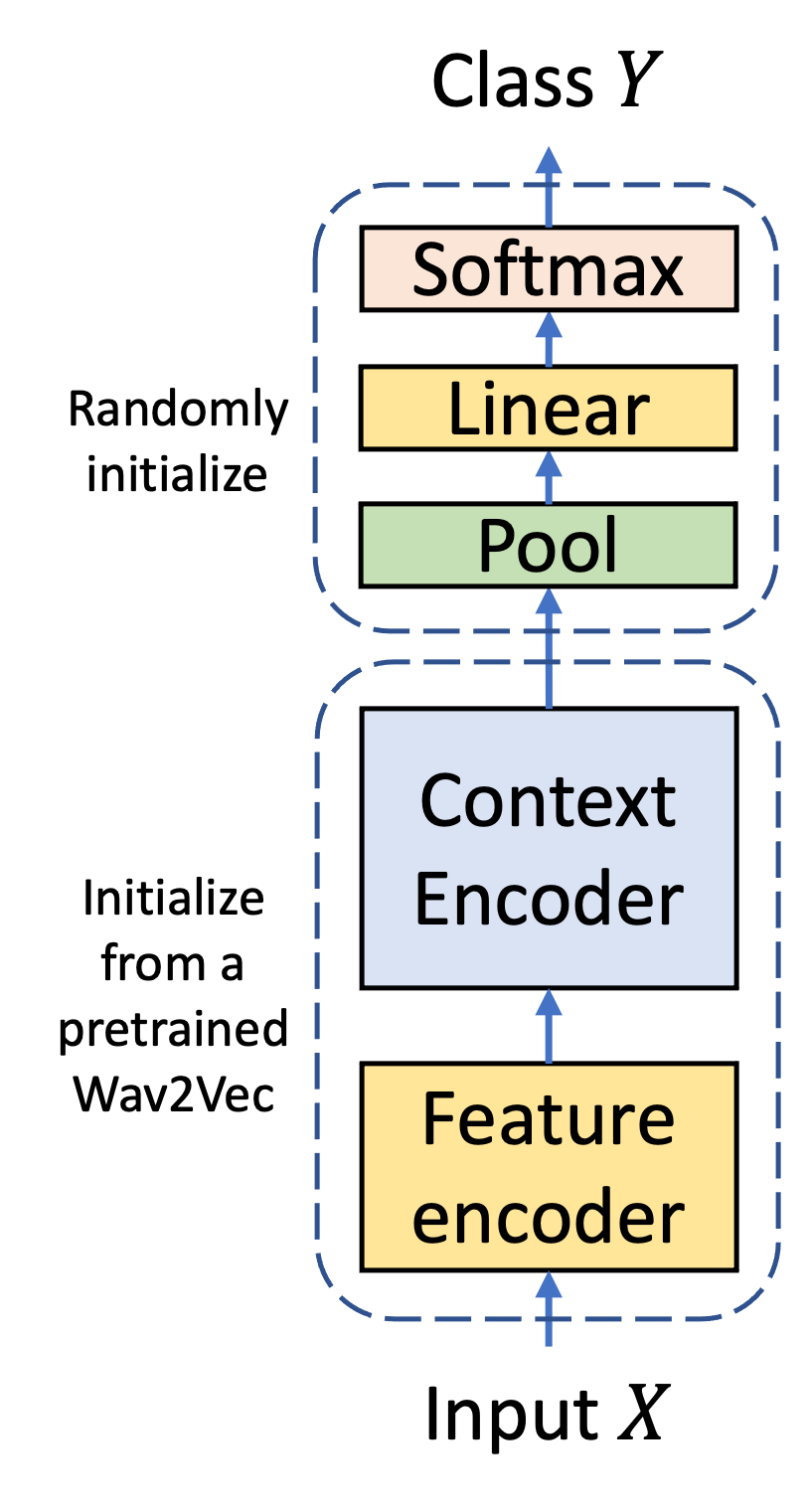}
\caption{A Wav2Vec encoder with pooling and softmax projection layer for utterance-level language id (LID) classification.}
\label{fig:wav2vec_lid}
\end{figure}
We illustrate our LID classifier architecture in Figure~\ref{fig:wav2vec_lid}. After we pre-training a log-mel wav2Vec model, we take the latest checkpoint and use it to initialize the bottom part of an LID classifier. The context representations $C = [c_1,...,c_T] \in \bR^{T\times D}$ are summarized by adding a pooling function ${pool}(\cdot): \bR^{T\times D} \xrightarrow{} \bR^{D}$. We explore several pooling operations such as:
\begin{enumerate} 
\item Mean pooling: $o = \sum_{t=1}^{T} {c_{t}}/{T}$.
\item Standard deviation pooling: $o = \sqrt{\sum_{t=1}^{T} {(c_t - \mu)}^2 / T}$


\item Self-attention pooling \cite{lin2017attentive}:
\begin{align}
    a &= \texttt{softmax}(w_{2}\texttt{GELU}(W_{1}c^{\intercal})) \in \bR^{T} \\
    o &= \sum_{t=0}^{T} a_t c_t \in \bR^{D}
\end{align} where $W_1 \in \bR^{U \times D}, w_2 \in \bR^{U}$ and $o$ is a weighted sum from $C$ based on attention vector $a$.
\item CLS-token pooling \cite{devlin-2019-bert}: Adding a special \texttt{[CLS]} token in the beginning of sequence and set $o=c_{1}$.
\end{enumerate}
After the pooling layer, we append a randomly initialized linear layer with $L$ output dimensions on top of a pooled representation $o$, and normalize it via a softmax function. We minimize the cross-entropy loss $\mathcal{L}=-\sum_{l=1}^{L} y_{l} log(p_{l})$ where $p_{l}$ is the predicted probability of speech utterance $X$ belonging to language $l$ and $y_{l} = 1$ if $l$ is the target class, otherwise $y_{l} = 0$.

\section{Experimental setup}
\subsection{Dataset}

We conducted the experiments on our in-house datasets. The training data consists of de-identified public videos with no personally identifiable information (PII), from where only the audio part is used. The dataset contains a fair amount of accented speech utterances and a good amount of local dialects for languages that are spoken across the globe. \atj{To pre-train an XLSR, we gathered up to 6.3 million hours of unlabeled audio. 
To pre-train a wav2vec-En, we required a large dataset with only English utterances, however, since we do not have language labels for the unlabeled video dataset, we created a `pseudo-en' dataset by using an in-house LID model to predict the class for each unlabeled audio and pick each utterance where the predicted class is 'en'.  Note, this model mostly serves as a comparison to cross-lingual pre-training with XLSR.}

For the input features, we extract 80-dimensions log-mel spectrogram with 25 milliseconds window size and 10 milliseconds step size. We normalize the value for each feature dimension by subtracting it with the mean and divide by the standard deviation. The mean and standard deviation are calculated from a random subset of the pre-training dataset.

\subsection{Pre-training setup}
Here, we describe each module configuration inside our pre-trained wav2vec model.
Inside a feature encoder, we have:
\begin{itemize}[itemsep=0em]
    \item Time-stride layer: reduce input sequence length by $R=4$ times (80 input dimension,  320 output dimension).
    \item Linear layer: 320 input dimension, 512 output dimension. 
\end{itemize}
Inside a context encoder, we have:
\begin{itemize}[itemsep=0em]
    \item Linear layer: 512 input dimension, 1024 output dimension.
    \item 1D Convolution layer: 1024 input dimension, 1024 output dimension, kernel size 48, filter groups 16.
    \item Transformers: 24 layers, 1024 input dimension, 16 attention head, 4096 feedforward dimension, GELU activation function, pre-layer norm \cite{xiong2020layer}.
    \item Linear layer: 1024 input dimension, 768 output dimension.
\end{itemize}
To generate quantized target $Q$ for contrastive task, we feed latent speech representation $Z$ into a quantization block with:
\begin{itemize}[itemsep=0em]
    \item Linear layer: 512 input dimension, 768 output dimensions.
    \item Gumbel VQ: 320 codebooks, 2 groups.
    \item Linear layer: 768 input and output dimension.
\end{itemize}
For masking over the latent speech representation $Z$, we sample $p=0.065$ as the starting indices and we mask the next $M=5$ frames. 
Overall, this model has 300 million parameters. 

We pre-trained two models with the same architecture but different inputs: \begin{enumerate} 
\item wav2vec 2.0 En is trained on English only.
\item XLSR is trained with all unlabeled audios. 
\end{enumerate}
We set the diversity loss hyperparameter $\lambda = 0.1$ for all experiments.
All models are trained using the Adam optimizer \cite{kingma2014adam} with learning rate $lr=1e-3$ for wav2vec 2.0 En and $lr=5e-3$ for XLSR up to 300000 updates. We also add weight decay $1e-2$ with $\ell^2$ weight penalty. We anneal the learning rate by using a linear decay learning schedule, with warm-up step up to 32000 updates and linearly decay to 0 after that. We crop the input sequence length up to 2000 frames (equals to 20 seconds). 

\subsection{Fine-tuning setup}
In the finetuning step, we randomly crop the audio into a 6 seconds chunk and extract 80-dimensions log mel spectrogram. On top of a wav2vec encoder, we added a pooling layer and a linear layer with $L$ output dimension. We prepare the finetuning dataset with 26 languages: English (en), Spanish (es), Arabic (ar), Indonesian (id), Vietnamese (vi), Portuguese (pt), Thai (th), Hindi (hi), Italian (it), French (fr), Turkish (tr), Tagalog (tl), Urdu (ur), German (de), Chinese (zh), Malayalam (ml), Bengali (bn), Russian (ru), Burmese (my), Malay (ms), Tamil (ta), Marathi (mr), Kannada (kn), Sinhalese (si), Japanese (ja), Dutch (nl). We prepare different amounts of supervised data per language: 10 minutes, 1 hour, 10 hours, 100 hours. \atj{This demonstrates the possibility of training LID on low resource scenario and improving LID result on high resource scenario on top of a self-supervised pre-trained model}. All finetuning models are trained with Adam optimizer with learning rate $lr=1e-4$, tri-stage learning rate schedule (10\% warm-up step, 40\% stay-step and 50\% decay step), weight decay $1e-2$ with $\ell^2$ weight penalty.
\atj{We also have several finetuning scenarios such as from scratch (without any pre-trained stage), from a monolingual wav2vec 2.0 En checkpoint, and an XLSR checkpoint. All models have the same architecture and number of parameters. 

During the finetuning training stage, we randomly sample 6 seconds audio chunks for each utterance. During the evaluation stage, we run the LID classifier on 6 seconds window and 3 seconds step size, average the language probability across multiple predictions and pick the highest probability as the prediction.}

\section{Results}

\begin{table}[t]
\begin{center}
\begin{tabular}{l|l|r|r|r|r}
    \toprule
\textbf{Lbl. } & \textbf{Pre-} & \multicolumn{4}{c}{\textbf{Test Accuracy (\%)}}                                                                                                                       \\ \cline{3-6} 
\textbf{/ lang}    & \textbf{training}& \textbf{0-6s} & \textbf{6-18s} & \textbf{18-$\infty$s} & \textbf{Overall} \\ \midrule
10 min  & None        & 7.1   & 9.5   & 10.6  & 9.6  \\
        & w2v2 En      & 71.3  & 73.1  & 76.1  & 74.2 \\
        & XLSR        & 85.4  & 88.8  & 90.8  & \textbf{89.2} \\ \hline
1 hour     & None        & 20.2  & 25.2  & 29.5  & 26.5 \\
        & w2v2 En      & 79.3  & 85.9  & 89.3  & 86.5 \\
        & XLSR        & 87.2  & 92.5  & 94.8  & \textbf{92.8} \\ \hline
10 hours    & None        & 48.3  & 61.9  & 71.8  & 64.5 \\
        & w2v2 En      & 86.8  & 93.3  & 95.6  & 93.4 \\
        & XLSR        & 88.2  & 94.3  & 96.1  & \textbf{94.2} \\ \hline
100 hours   & None        & 72.2  & 84.9  & 90.7  & 86.7 \\
        & w2v2 En      & 89.5  & 95.7  & 97.3  & 95.5 \\
        & XLSR        & 90.3  & 95.9  & 97.2  & \textbf{95.7} \\
    \bottomrule
\end{tabular}
\caption{LID test accuracy for 26 languages setup using different amounts of labeled training data per language (10 minutes - 100 h). 
We compare three scenarios: training an LID model without pre-training (None), fine-tuning a monolingually pre-trained wav2vec 2.0 model (w2v2 En), and fine-tuning a cross-lingually pre-trained model (XLSR). 
}
\label{tbl:main_exp_26}
\end{center}
\end{table}

\subsection{Language identification for 26 languages}
The ability to train LID models with very little labeled data is important in order to be able to extend speech technology to the thousands of languages and dialects spoken around the world.
The test accuracy for 26 languages experiments are calculated from a test set that contains a total of 3700 hours from 26 languages. The accuracy are shown in Table~\ref{tbl:main_exp_26}. We calculated the accuracy for short utterances (shorter than 6 seconds), medium utterances (6-18 seconds), long utterances (longer than 18 seconds), and overall accuracy.
Based on our experiment, we show that training from scratch (None) performs particularly poorly with just 10 minutes of labeled data. 
On the other hand, XLSR achieves over 89.2\% accuracy.
Pre-training on more languages performs better with little labeled data and in the high-resource labeled data regime, the labeled data provides sufficient learning signal.
There is a similar trend of training from scratch (no pre-training) improving with more labeled data but even with large amounts of labeled data, there is still a sizeable gap to pre-trained models.

\begin{figure}[t]
\centering
\includegraphics[width=1.0\linewidth]{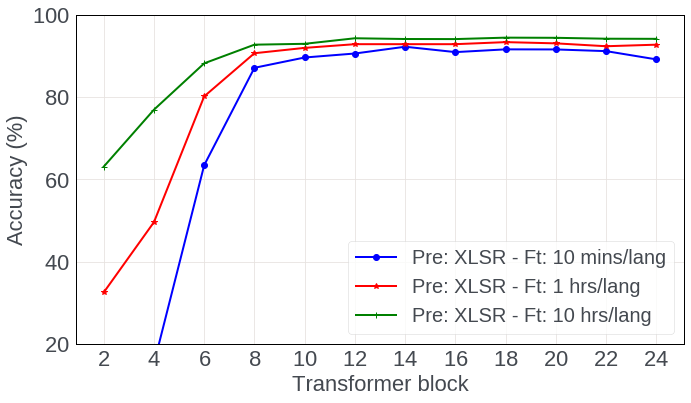}
\caption{LID overall accuracy when using the output of different Transformer blocks from a pre-trained (pre) XLSR model as input to the LID classifier. Accuracy is in terms of the 26 language setup and using 10 minutes, 1 and 10 hours of labeled data per language for fine-tuning (ft).}
\label{fig:probe_exp}
\end{figure}

\subsection{Ablations}

Next, we explore the effect of using representations from different Transformer blocks of the pre-trained models. For this section, we focus our experiment on 26 language setup using 10 minutes, 1 and 10 hours of labeled data per language for fine-tuning.
Figure~\ref{fig:probe_exp} shows that the middle and upper parts of the pre-trained model (from 8th to 24th layer) perform significantly better than the lower part (from 2nd to 6th layer). 
The result suggests that we could prune up to $2/3$ of the context encoder blocks, which reduces the time and memory usage during fine-tuning and inference while maintaining good accuracy. 
Additionally, by keeping only eight Transformer blocks, we reduce the number of parameters from 300 million down to 100 million.

\begin{table}[t]
\begin{center}
\begin{tabular}{l|r|r|r|r}
    \toprule
    \textbf{Aggregation}  & \multicolumn{4}{c}{\textbf{Accuracy (\%)}} \\
    \cline{2-5}
    \textbf{strategy} & \textbf{0-6s} & \textbf{6-18s} & \textbf{18-$\infty$s} & \textbf{Overall} \\
    \midrule
    Max             & 86.6 & 92.7 & 94.8 & 92.8 \\
    Mean+Max+Min    & 88.1 & 92.9 & 94.7 & 93.0 \\
    Mean+Max        & 88.5 & 93.1 & 94.8 & 93.2 \\
    Mean+Std        & 84.2 & 90.9 & 93.4 & 91.1 \\
    \texttt{[CLS]} Token      & 85.4 & 91.4 & 93.9 & 91.7 \\
    Self Attention & 87.0 & 92.0 & 94.1 & 92.2 \\
    \bottomrule
\end{tabular}
\caption{LID accuracy for different strategies to aggregate the context representations of an XLSR model on the 26 language setup using 1 hour of labeled data per language for fine-tuning.}
\label{tbl:pooling_exp}
\end{center}
\end{table}

So far we used mean pooling to aggregate the output of the pre-trained models for a given speech utterance into a single vector.
Table~\ref{tbl:pooling_exp} compares this strategy to max pooling, concatenated mean+max+min pooling, concatenated mean+max pooling, concatenated mean+std (statistical pooling \cite{snyder2017deep}), first-step class tokens (\texttt{[CLS]}) \cite{devlin-2019-bert}, and self-attention pooling \cite{lin2017attentive}. 
The result suggests that mean pooling works very well compared to the alternatives and provides a simple way to aggregate the context information.

\section{Conclusion}

In this paper, we demonstrated the benefit of using self-supervised pre-trained representations learned on unlabeled speech data to improve language identification. 
We showed that pre-training is more effective than training LID models solely from labeled data and cross-lingual representations are particularly effective for low-resource setups, where little labeled data is available. 
This is important to enabling speech technology for many more languages spoken around the world.
Using only 10 minutes of labeled data per language, a cross-lingually pre-trained LID can achieve an accuracy of over 89.2\% on a 26 language setup. Additionaly, we also observe the benefits of cross-lingual pre-training on LID with higher amount of labeled data.
We find that we can prune up to two-thirds of the pre-trained model while achieving the same accuracy. For future work, we may explore how to make these models more efficient for inference since pre-trained models are still very large.

\vfill\pagebreak

\label{sec:refs}

\bibliographystyle{IEEEbib}
\bibliography{strings,refs}

\end{document}